\documentclass[10pt,twocolumn,letterpaper]{article}

\usepackage{cvpr}
\usepackage{times}
\usepackage{epsfig}
\usepackage{graphicx}
\usepackage{amsmath}
\usepackage{amssymb}
\usepackage{booktabs}
\usepackage{multirow}
\usepackage[numbers,sort]{natbib}
\usepackage[american]{babel}

\def\RR{{\mathbb R}}    
 
\def\EE{{\mathbb E}}    
\def\11{{\mathbf 1}}    

   \def\bM{{\mathbf M}}                              \def\b1{{\mathbf 1}}

\def\cA{{\mathcal A}}               \def\cD{{\mathcal D}}              \def\cX{{\mathcal X}} \def\cY{{\mathcal Y}}  

\newcommand{\Mult}[1]{\operatorname{Mult}({#1})}

\usepackage[pagebackref=true,breaklinks=true,letterpaper=true,colorlinks,bookmarks=false]{hyperref}

 \cvprfinalcopy 

\def\cvprPaperID{8540} 

\ifcvprfinal\fi
\begin{document}

\title{Mixup Regularization for Region Proposal based Object Detectors}

\author{Shahine Bouabid\textsuperscript{1, 2, 3}\qquad Vincent Delaitre\textsuperscript{1}\\
\textsuperscript{1}Deepomatic\qquad \textsuperscript{2}CentraleSupelec\qquad \textsuperscript{3}ENS Paris-Saclay\\
{\tt\small shahine.bouabid@student.ecp.fr, vincent@deepomatic.com}
}

\maketitle

\begin{abstract}
Mixup | a neural network regularization technique based on linear interpolation of labeled
sample pairs | has stood out by its capacity to improve model’s robustness and generalizability
through a surprisingly simple formalism. However, its extension to the field of object detection
remains unclear as the interpolation of bounding boxes cannot be naively defined. In this paper, we
propose to leverage the inherent region mapping structure of anchors to introduce a mixup-driven
training regularization for region proposal based object detectors. The proposed method is benchmarked
on standard datasets with challenging detection settings. Our experiments
show enhanced robustness to image alterations along with an ability to decontextualize detections, resulting in an improved generalization power.
\end{abstract}

\section{Introduction}
A major upturn in computer vision has come with the bloom of deep learning. In particular, convolutional
neural networks (CNNs)~\cite{NIPS2012_4824, VGG, resnet, xception, googlenet, mobilenet} have demonstrated their capacity to learn discriminative visual features, and have allowed tremendous progress in
tasks such as image classification.

The object detection task naturally extends image classification by introducing the concept of location. Given
an image, we aim at spatially delimiting all object instances of interest before classifying them among a finite
set of categories. The difficulty of such task arises from the diversity of possible viewpoints,
occlusions, positions and lighting conditions which challenges the robustness of proposed methods.
Deep learning-based object detectors~\cite{fasterrcnn, yolo, ssd} have greatly contributed to overcome these limitations and obtain state-of-the-art performance~\cite{DL_detection_review, recent_advances_dl}. They are consequently used in a wide range of real-world applications today.

\begin{figure}[t]
\begin{center}
   \includegraphics[width=\linewidth]{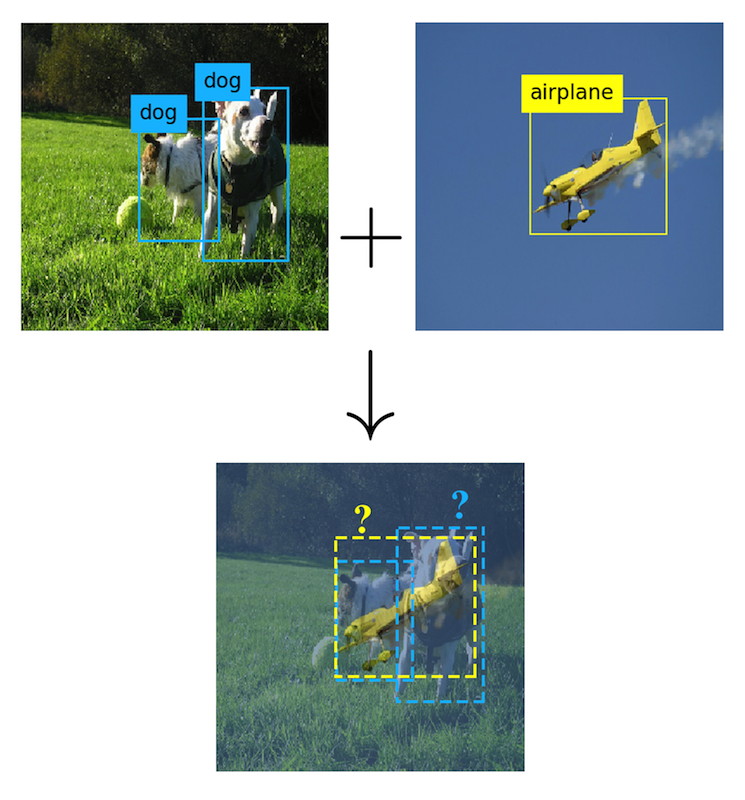}
\end{center}
   \caption{The mix between two sets of bounding boxes cannot be trivially defined, as their number,
   location and labels are not necessarily comparable.}
\label{fig:not_trivial}
\end{figure}

Both classification and detection tasks are usually formulated as supervised learning problems where
the objective is to minimize the prediction error of the model on a dataset of fully annotated images.
The image distribution in the dataset may strongly affect the model performance \cite{CNN_imbalance, DL_imbalance}. High-density areas of the
dataset can be thought of as well-populated clusters in the image space, likely to
convey spatially uniform information and to produce consistent predictions of the model. On the contrary, low-density areas, \ie underrepresented
areas of the image space, are regions one has poor knowledge of. A classification decision boundary should intuitively lie in a low-density area, away
from dense clusters~\cite{regularization_taxonomy}. But it should also tread lightly in such area, in accordance with the lack of data. Bafflingly, deep
classification networks tend
on the contrary
to present irregularly sharp decision boundaries, which are dangerously close to high-density
regions~\cite{ICT, manifold_mixup}.

Adversarial attacks are maliciously crafted signal perturbation which can purposely sway the output of the network while being imperceptible to humans ~\cite{goodfellow_adversarial}. They point out how skillfully designed noise can exploit the model's lack of regularity, even at a pixel-wise scale.

In this regard, object detection flaws are dual~\cite{toward_robust_detection}: not only classification
but also localization can suffer from an untrustworthy learned distribution. This is particularly critical in
applications such as self-driving cars or surveillance. Beyond image-level adversarial attacks~\cite{transferable_attacks, adversarial_semantic}, detectors are also vulnerable to transplantation of image parts, referred to as \emph{patching}. They can trigger misdetection, even away from the patch.
The patch can be tricked to interfere with the prediction~\cite{dpatch, physical_patch}, but more
interestingly, even naive patching has been found to harm detectors~\cite{elephant}. Furthermore, the latter has inspired several data augmentation
strategies~\cite{compositing, cordelia, cut_paste,  indoor_scenes}.

The work by Zhang~\etal~\cite{mixup} introduces a training regularization method called \textit{mixup}.
It extends the representation of the dataset by linearly interpolating samples along with their labels.
Doing so, one is actually able to mitigate the neural network's degree of extrapolation with respect to our actual knowledge of data in the low-density regions of the dataset
with elegant simplicity and no additional inference time. It has already proven its ability to improve the
model's robustness and generalization power on a colorful range of tasks such as supervised and
semi-supervised image classification~\cite{mixup, mixmatch}, text and acoustic scene classification
\cite{mixup_text, mixup_acoustic} as well as medical image segmentation~\cite{mixup_medical}. However,
while mixing up labels comes in naturally through the convex combination of their one-hot
representation, this operation is less obvious when it comes to bounding boxes as depicted in Figure~\ref{fig:not_trivial}.

In region-based detectors~\cite{fasterrcnn, yolo, ssd}, a predefined set of boxes called \textit{anchors} constitutes a comprehensive tiling of possible object locations over the image. The role of the neural
network is then to provide a classification score and coordinates regression over each anchor. Leveraging
the comprehensiveness assumption of such regions mapping, we propose in this paper a simple and
general method to perform mixup-regularized training on region proposal based object detectors.
Consistently with the behavior of mixup-trained classification networks, we find the proposed solution to
outperform baseline
models on Pascal VOC~\cite{pascal_voc} and MS COCO~\cite{coco} datasets, but most importantly to demonstrate evidence of enhanced robustness.

The contributions of this paper are threefold: \textbf{(i)} we provide insights on mixup-based training
approaches to highlight how object detectors could benefit from them; \textbf{(ii)} we propose a general
method to adapt mixup to region proposal based detectors; \textbf{(iii)} we verify the proposed strategy
effectiveness with extensive experiments on the Single Shot Multibox Detector (SSD) meta-architecture~\cite{ssd} and show extended generalization capabilities.

\section{Related work}

\subsection{Mixup Regularized Training for Classification}

In the case of a supervised classification problem, let $x, x'$ be two independent data samples labeled
with $y, y'$, respectively. Let
$\lambda\in[0,1]$ be a mixing weight and $\tilde x=\lambda x + (1 - \lambda)x'$ the mixed sample with
label $\tilde y$. While $\tilde y$ could be any combination of $y$ and $y'$, in the interest of a regularized learning
at $\tilde x$, the natural choice of setting $\tilde y = \lambda y + (1 - \lambda) y'$ appears to be
supported by several observations. Firstly, according to Occam's razor~\cite{mixup}, without further
constraints on $\tilde y$, one should aim for the plainest solution. This choice also smoothens the model
by fostering a linear behavior in-between training samples, consistently with the \emph{smoothness
assumption}~\cite{Chapelle}. Finally, from an information theory perspective, we have by linearity of the expectancy:
\begin{align}
H(\tilde y) & = \EE_{\lambda y + (1 - \lambda)y'}[-\log\tilde y]\\
& = \lambda H(y, \tilde y) + (1 - \lambda)H(y', \tilde y)
\end{align}
$H$ denoting entropy or cross-entropy when appropriate.
The amount of information endowed to $\tilde y$ is hence exactly a convex combination of
\textit{``what do we know about $\tilde y$ from the perspective of $y$ and $y'$''}.

\begin{figure}[h!]
\centering
   \includegraphics[width=\linewidth]{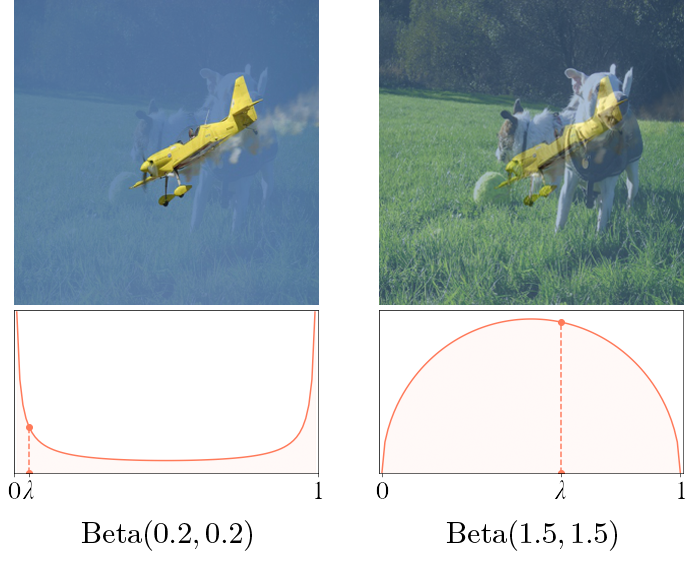}
   \caption{Visual comparison of different random mixing weight distributions. On the left, for $\alpha < 1$, the probability density function is such that most drawn $\lambda$ are close to $0$ or $1$, resulting in a lightly noised image. On the right, $\alpha > 1$ gives a probability density function such that most random mixing weights $\lambda$ are near $0.5$, producing a mixed image close to the "average" image, allowing both images to be effectively visible.}
\label{alpha_influence}
\end{figure}

A typical choice is to let the mixing ratio $\lambda$ follow a $\operatorname{Beta}(\alpha, \alpha)$ distribution
with $\alpha > 0$. This distribution can be thought of as a prior over a fair heads or tails game.
As depicted in Figure \ref{alpha_influence}, the greater alpha is, the more blended pairs of images are.

Mixup hence introduces very little computational cost and is shown to consistently improve image
classification models across datasets. It favors robustness by, on the one hand, leading to smoother
decision boundaries lying in low-density areas~\cite{mixup}, and on the other hand, enabling
compression of class-specific representations~\cite{manifold_mixup}.

\subsection{Object Detection and Mixup}

\begin{figure*}[t]
\centering
\includegraphics[width=\linewidth]{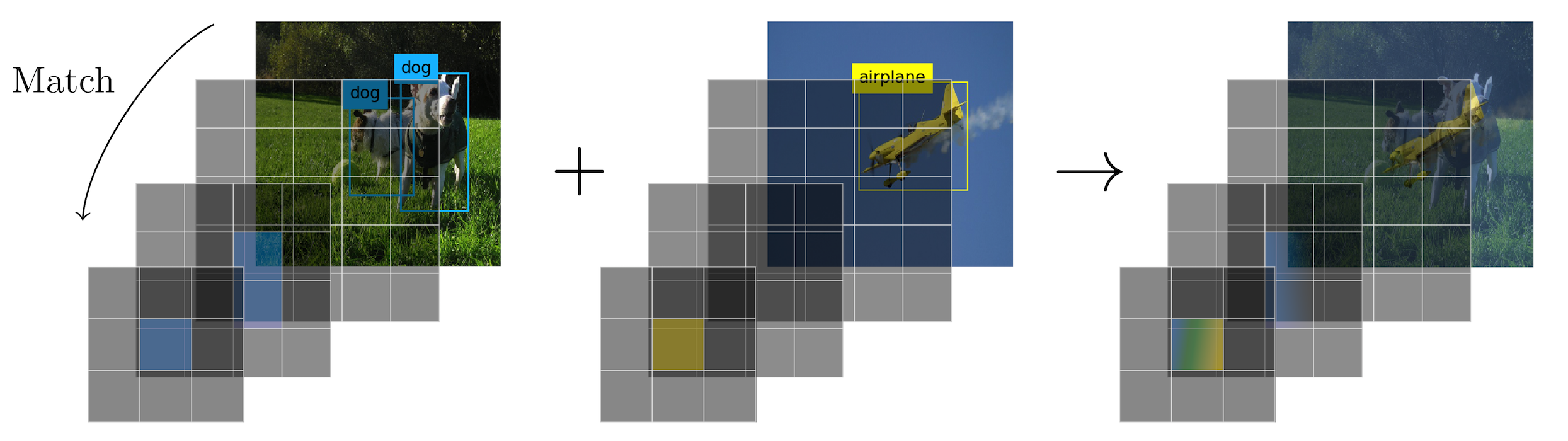}
\caption{Our method for mixing bounding boxes. Ground truth bounding boxes are matched to multiscale anchors meeting a sufficient overlap
criterion. Three illustrative anchor grids are represented, the smallest corresponding to greater scale
anchors. Dark cells are left unmatched and assigned the background label, blue cells are matched with
\emph{dog} bounding boxes and yellow cells to the \emph{airplane} bounding box. Images are mixed
pixel-wise and anchor grids blended anchor-wise to obtain the mixed image anchor representation.}
\label{anchormix}
\vspace*{-0.8em}
\end{figure*}

Zhang~\etal~\cite{bag_of_freebies} propose a visually coherent mixing strategy for object detection.
Mixed images are obtained by geometry preserving alignment and standard pixel-wise interpolation.
Bounding boxes are simply stacked to simulate object's co-occurrence and help the model deal with
complex settings such as occlusion and image perturbation. The mixing ratio hyperparameter $\alpha$ is
set to be greater than 1 to obtain a mixed image visually consistent with all bounding boxes.

Motivated by Rosenfeld \etal.~\cite{elephant}, they demonstrate the effectiveness of their approach on
the YOLOv3~\cite{yolov3} model by trying to detect a transplanted elephant patch on an indoor image. While
their baseline model struggles due to a challenging and incongruous setting for an elephant,
the mix-trained model appears to be less affected by the context. It shows greater recall on the elephant
and reinforced robustness to this perturbation.

However, whereas a linear relationship holds in mixup between the input and the supervision signal, the
mixed set of bounding boxes used in~\cite{bag_of_freebies} is independent of the mixing ratio. Naively stacking bounding boxes
hence allegedly strips the mixup procedure from enforcing smoothness.

\section{Method}

Let $\cX, \cY$ respectively denote the samples and targets space, where $\cX\subseteq\RR^d$, and
$\cD\in(\cX\times\cY)^n$ an identically and independently distributed dataset inducing an empirical joint
probability distribution $P_\cD$ over $\cX\times\cY$. Let $f_\theta = \cX \rightarrow\cY$ be our
parametrized neural network hypothesis and $\ell$ some non-negative differentiable training criterion.

When addressing a classification problem over $C$ possible classes, the label space is
$\cY = \Mult{C}$, where $\Mult C$ denotes multinomial distributions over $\{1, \ldots, C\}$, and $\ell$
would typically be a cross-entropy.
As proposed by Zhang~\etal~\cite{mixup}, the mixup classification
problem simplifies to the minimization of the expected risk:
\begin{equation}\label{cls_expected_risk}
R(\theta) = \EE[\ell(f_\theta(\lambda x + (1 - \lambda) x'), \lambda y + (1 - \lambda) y')],
\end{equation}
where $(x, y), (x', y')\sim P_\cD$ and $\lambda\sim\operatorname{Beta}(\alpha, \alpha)$.

When it comes to object detection, each element of $\cY$ is composed of one or multiple labeled boxes
and one cannot naively interpolate different sets of bounding boxes as the latter are not trivially
comparable.

Region proposal based object detectors such as Faster-RCNN~\cite{fasterrcnn}, SSD~\cite{ssd} or
YOLO~\cite{yolo} allow a CNN to cast the detection problem as multiple classification and
regression problems over a set of predefined regions called \emph{anchors}~\cite{fasterrcnn}. Those
anchors are meant to constitute a comprehensive multi-scale discrete tiling of possible object
locations across the image. A first part of the model, called \emph{region proposal network} (RPN), predicts a probability distribution over classes | 
augmented with the background class | and four refining
coordinates offsets for each anchor.

Each anchor being defined by its center coordinates and dimensions, let $\cA\subset\RR^4$ denote this finite prior anchor set. We introduce the \emph{anchor representation space}:

\begin{equation}
\Gamma(\cA) = (\RR^4_\bullet\times\Mult C)^{|\cA|},
\end{equation}
where $\Mult C$ stands for classification distributions over background and object labels, and $\RR^4_\bullet = \RR^4\cup\{\varnothing\}$ stands for offsets between box coordinates. The symbol $\varnothing$ represents an undefined offset, as explained below.

The RPN is here an application $f_\theta = \cX\rightarrow\Gamma(\cA)$ predicting over each anchor $a\in\cA$ a coordinates offset vector $\hat
\delta_a\in\RR^4$ and a label $\hat p_a\in\Mult C$.

We also define a \emph{matching strategy} to
associate ground truth bounding boxes to one or more of these anchors, thus enabling comparison with
the prediction. This matching strategy standardly uses the Intersection-over-Union (IoU) kernel to
compare bounding boxes given a similarity threshold. Formally, we assign each anchor $a\in\cA$ a
ground truth offset vector $\delta_a\in\RR^4_\bullet$ and a label $p_a\in\Mult C$. If $a$ is matched to some
ground truth box $b$, \ie if $\operatorname{IoU}(a, b)$ is greater than a given threshold $\tau$, we set $\delta_a$ to the coordinates difference between $a$ and $b$ and $p_a$ to the ground truth label.
Otherwise, $p_a$ is set to the background label and $\delta_a=\varnothing$.

By using this \emph{matching strategy}, we can translate ground truth bounding boxes $y \in \cY$ into a family $\{(\delta_a, p_a)\}_{a\in\cA}\in\Gamma(\cA)$
where each anchor is tied to its own probability distribution and regression
coordinates. We embody it by the matching operator $\bM = \cY\rightarrow\Gamma(\cA)$.

Assuming the comprehensiveness of regions coverage
by anchors, $\Gamma(\cA)$ constitutes a proper representation space for bounding boxes, embedding
both localization and classification information. Driven by this observation, we propose the following
mixing strategy depicted in Figure \ref{anchormix}.

We consider a pair of images $x, x'$ with respective sets of bounding boxes $y, y'$, and a
random mixing weight $\lambda\sim\operatorname{Beta}(\alpha, \alpha)$. As usual, we use standard
linear interpolation to define the mixed image and predict:
\begin{equation}
f_\theta(\lambda x  + (1 - \lambda) x') = \{(\hat\delta_a, \hat p_a)\}_{a\in\cA}.
\end{equation}

However, instead of naively mixing $y$ and $y'$, we propose to mix their matched anchor representations, namely $\bM(y)
= \{(\delta_a, p_a)\}_{a\in\cA}$ and $\bM(y') = \{(\delta'_a, p'_a)\}_{a\in\cA}$.
To do that, let's define by $\operatorname{BoxMix}_\lambda$ the desired stochastic mixing operator such that:
\begin{equation}
\operatorname{BoxMix}_\lambda(y, y') = \{(\tilde\delta_a, \tilde p_a)\}_{a\in\cA}.
\end{equation}

As it does intuitively make little sense to mix offsets, we decide to simply
define $\tilde\delta_a$ as the offset from the dominant image, \ie with greater mixing ratio, as:
\begin{equation}
\tilde \delta_a= \begin{cases}
\delta_a & \text{if } \lambda > \frac{1}{2},\\
\delta'_a & \text{otherwise.}
\end{cases}
\end{equation}

On the other hand, inspired by classification mixup, we propose to perform linear interpolation on
anchor labels and set:
\begin{equation}
\tilde p_a = \lambda p_a + (1-\lambda)p'_a
\end{equation}

By reproducing classification mixup anchor-wise, we thus build an explicit interpolation map of each
singular anchor probability distribution.

The latter can eventually be fed to the RPN standard robust regression and classification losses as:
\begin{align}
L_\text{reg}(\tilde\delta_a, \hat\delta_a) & = \11_{\{\tilde\delta_a\neq\varnothing\}}\operatorname{smooth}_{L_1}(\tilde\delta_a - \hat\delta_a) \\
L_\text{cls}(\tilde p_a, \hat p_a) & = \EE_{\tilde p_a}[-\log \hat p_a]
\end{align}

Drawing a parallel with Equation~\ref{cls_expected_risk}, we formulate the region
based object detection mixup problem as the minimization of the following expected risk:
\begin{equation}
R(\theta) = \EE[\ell( f_\theta(\lambda x + (1 - \lambda)x'), \operatorname{BoxMix}_\lambda(y, y'))],
\end{equation}
where $\ell$ accounts for the anchors sampling strategy and both regression and classification losses.

\section{Experiments}

\begin{figure*}[t]
\centering
   \includegraphics[width=\linewidth]{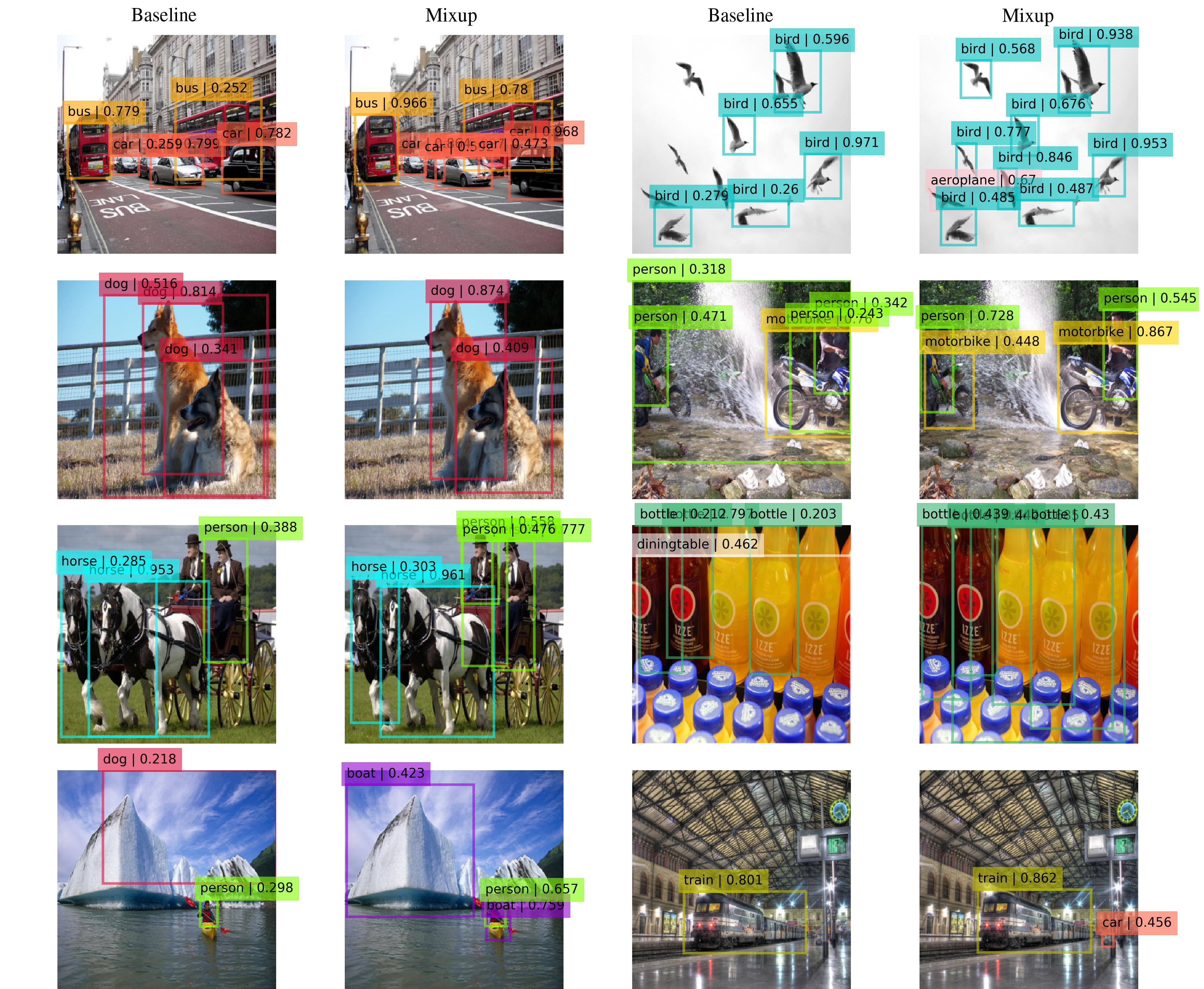}
   \caption{Detection examples from VOC07 test set, best viewed in color. 1\textsuperscript{st} row: mixup enhances recall and confidence but yields false positives. For example, a bird is confused for an airplane. 2\textsuperscript{nd}-3\textsuperscript{rd} row: mixup helps with occlusion and detection of multiple contingent objects. 4\textsuperscript{th} row: example of greater confidence over small objects detection, at risk of degrading precision.}
\label{florilege}
\end{figure*}

\begin{table*}[h]
\centering
\footnotesize
\setlength\tabcolsep{2pt}
\begin{tabular}{clccccccccccccccccccccc}\toprule
 Train set & Method   & mAP   & aero   & bike  & bird  & boat  & bottle & bus   & car   & cat   & chair & cow   & table & dog   & horse & mbike & person
 & plant & sheep & sofa & train & tv \\\toprule

\multirow{3}{*}{\begin{tabular}[c]{@{}l@{}}07\end{tabular}} &  Baseline & 66.2\scriptsize{$\pm$0.4} & 81.9  & \textbf{77.8} & 81.4 & \textbf{66.5} &
50.3  & 55.5 & 75.9 & 72.6 & 51.4 & 63.3 & \textbf{40.0} & 65.8 & 74.6 & \textbf{66.7} & 81.1  & 60.1 & 75.4 & 42.8 & 77.6 & 63.8 \\

  & BoxStack~\cite{bag_of_freebies}   & 62.1\scriptsize{$\pm$0.7} & 80.8  & 75.7 & 77.0 & 55.5  & 51.4 & 53.0 & 76.5 & 65.0 & 49.9 & 55.1 & 29.5 & 55.3 & 73.2 & 65.2  & 79.2 & 56.6 & 67.4 & 42.7 & 73.2 & 59.9\\

  & Mixup   & \textbf{68.0\scriptsize{$\pm$0.1}} & \textbf{83.6}  & 77.2 & \textbf{82.8} & 65.6 & \textbf{52.2}  & \textbf{59.3} &
  \textbf{77.5} & \textbf{76.5} & \textbf{54.8} & \textbf{67.1} & 38.8 & \textbf{67.7} & \textbf{77.5} & 65.0 & \textbf{81.7}  & \textbf{62.6} &
  \textbf{78.0} & \textbf{47.5} & \textbf{80.1} & \textbf{64.9} \\\midrule

\multirow{3}{*}{\begin{tabular}[c]{@{}l@{}}07+12\end{tabular}} &  Baseline       & 72.5\scriptsize{$\pm$0.1} & 85.5 & 81.9 & 86.4 & 72.7 &
\textbf{54.0}  & 65.8 & 80.5 & 79.6 & 59.2 & 72.6 & \textbf{49.7} & 73.7 & 77.1 & \textbf{74.3} & 84.4  & \textbf{65.8} & \textbf{81.1} &
\textbf{55.8} & 82.7 & 66.5\\

& BoxStack~\cite{bag_of_freebies}   & 68.2\scriptsize{$\pm$0.5} & 83.8  & 79.5 & 83.6 & 65.8 & 53.1  & 63.0 & 79.2 & 73.4 & 56.9 & 65.7 & 39.6 & 65.6 & 73.7 & 73.7 & 81.8  & 61.3 & 75.0 & 48.4 & 78.9 & 64.1 \\

& Mixup & \textbf{73.1\scriptsize{$\pm$0.2}} & \textbf{87.2}  & \textbf{82.6} & \textbf{87.7} & \textbf{74.5} & 53.7  & \textbf{66.6} &
\textbf{81.3} & \textbf{81.4} & \textbf{60.4} & \textbf{75.5} & 47.2 & \textbf{74.5} & \textbf{77.5} & 72.5 & \textbf{84.6}  & 64.8 & 81.0 & 55.4 &
\textbf{83.8} & \textbf{68.0}\\\toprule

\end{tabular}
\caption{PASCAL VOC07 \texttt{test} detection results; "Baseline": standard SSD training; "BoxStack": our re-implementation of bounding boxes stacking from~\cite{bag_of_freebies}; "Mixup": our object detection strategy. The results are averaged over 5 independently trained models.}
\label{VOC_results}
\end{table*}

\begin{table*}[h!]
\centering
\small
\begin{tabular}{lcccccccccccc}\toprule
  Method      & AP\textsubscript{[0.5:0.95]} & AP\textsubscript{50} & AP\textsubscript{75} & AP\textsubscript{S} & AP\textsubscript{M} & AP\textsubscript{L} & AR\textsubscript{1} & AR\textsubscript{10} & AR\textsubscript{100} & AR\textsubscript{S} & AR\textsubscript{M} & AR\textsubscript{L} \\\toprule
Baseline      & \textbf{19.7}    & 33.7                                  & \textbf{20.3}                         & \textbf{3.7}                         & 20.4                                 & \textbf{34.6}                        & 19.2                                 & 26.2                                  & 26.9                                   & 5.0                                  & 27.1                                 & 47.4                                 \\
BoxStack~\cite{bag_of_freebies}      & 15.6                &       28.5                       & 15.4                             & 2.4                                    & 17.0                                    & 27.4                                  & 16.7                                  & 23.4                                 & 24.3                          & 3.7                                    & 24.8                                    & 41.9         \\
Mixup & 19.5             & \textbf{34.3}                         & 19.7                                  & 3.5                                  & \textbf{20.7}                        & 33.9                                 & \textbf{19.6}                        & \textbf{27.9}                         & \textbf{29.2}                          & \textbf{6.1}                         & \textbf{30.5}                        & \textbf{49.1}                        \\\toprule
\end{tabular}
\caption{MS-COCO \texttt{val-2017} detection results; Methods naming is similar to Table~\ref{VOC_results}.}
\label{COCO_results}
\end{table*}

\subsection{Experimental and Implementation Details}

\begin{figure*}[t]
\begin{center}
   \includegraphics[width=0.83\linewidth]{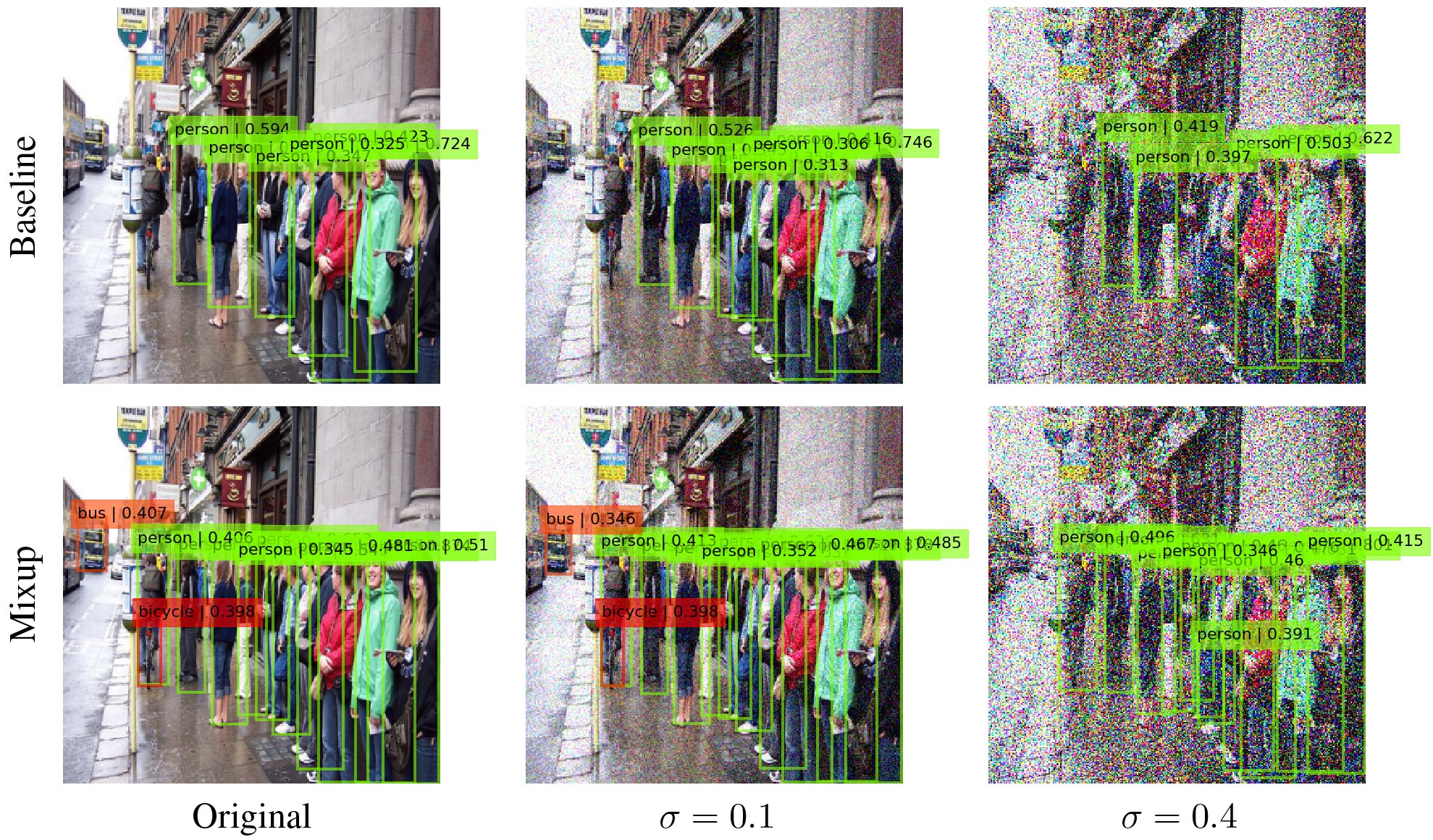}
\end{center}
   \caption{Example of detections on an image corrupted with Gaussian noise for baseline and mixup models. Predictions are initially more accurate for the mixup model with better recall. Low-corruption ($\sigma=0.1$) undermines predictions confidence without significantly degrading overall detections for both. Mixup is more resistant to greater corruption ($\sigma=0.4$), maintaining a reasonable recall although inaccurate detections start to appear.}
\label{noise_people}
\vspace*{-1.2em}
\end{figure*}

The SSD~\cite{ssd} meta-architecture is used in all experiments as a representative of region
proposal based detectors. We use an ImageNet~\cite{imagenet} pretrained ResNet-50~\cite{resnet}
backbone, whose pooling architecture naturally extracts multiscale feature representations.
We then map these feature representations to judiciously dimensioned anchors by following
recommendations of~\cite{github_ssd}.
As for the training criterion, we stick with Liu~\etal~\cite{ssd} original procedure and perform
hard negative mining to feed standard robust regression loss and classification cross-entropy.

We experiment on both PASCAL VOC~\cite{pascal_voc} and MS-COCO datasets~\cite{coco}. For VOC,
training is performed on standard 07 or 07+12 \texttt{trainval} union consisting of 16k images for 20
classes, and evaluate on the 5k images of the VOC07 \texttt{test} set. MS-COCO is split into 115k
images
(\texttt{train-2017}) for training and 5k images for validation (\texttt{val-2017}) with 80 object classes.
Reported results follow usual challenge conventions, and unless specified otherwise, mean Average Precision (mAP) is computed for 0.5~IoU
threshold.

As mixup endeavors lower gradient norms~\cite{mixup}, we opt for an adaptative optimizer
and train all models with Adam~\cite{adam}. A base learning rate of 3e-4, with 0.9 momentum and
5e-4 weight decay are used. The learning rate is decayed at the end of each epoch by factor 0.95. We use
a batch size of 32. All images are normalized, resized to 300$\times$300 and undergo the following data augmentation procedure: random
horizontal flips, color jittering and random cropping.

The \emph{matching strategy} threshold $\tau$ is set to 0.5 and each groundtruth box must be matched to at least one anchor.
Regarding post-processing, we use the default 0.45 non-maximal-suppression~\cite{nms} threshold as
suggested by Liu~\etal~\cite{ssd} and keep the top 200 detections at most. Visualizations are produced
by additionally discarding proposals with a score lower than some given threshold, typically 0.3.

Our code is implemented in PyTorch~\cite{pytorch} and ran on an Nvidia GeForce GTX 1080.

\subsection{Mixed Detection Results}

We compare the baseline SSD model against our mixup model and our re-implementation of naive bounding boxes stacking~\cite{bag_of_freebies}. We report results in Tables
\ref{VOC_results} and \ref{COCO_results}.
For fair comparison, the baseline model follows exactly the same implementation and training procedure as the proposed method. To comply with visual coherence, we use $\alpha=1.5$ in our
re-implementation of~\cite{bag_of_freebies}, but discard geometry preserving alignment and simply mix images by resizing them to the same size. For our model, we find $\alpha=0.2$ to give the desired
mixing ratio distribution. In practice, mixed images are hence simply noised as illustrated in Figure \ref{alpha_influence}. Yet, one could argue that
the mixup noise is crafted to follow a distribution proportional to the dataset empirical distribution. In the same way, adversarial attacks consist of
noises distorted to sway predictions.

Overall, as for classification, mixup consistently improves mAP. As shown in Figure \ref{florilege}, mixup helps the model dealing with object
multiplicity and contingency, leading thus to enhanced recall power. Moreover, whereas in classification mixup did soothe the model's over-confidence,
in our case it turns out to foster confident predictions. While confident detection is a desired behavior, it also tends to occasionally be counterproductive,
undermining its precision by clumsily "overachieving" detection. A judicious tuning of proposal filtering is thus required.

Although we believe to have rigorously reproduced the box stacking experiment of~\cite{bag_of_freebies} but for image alignment, we are surprised to notice it
actually drastically hurts performances. Overmixing samples and never showing realistic images to the model could hamper regularization by
introducing more confusion than it should. Indeed, in image classification, large values of $\alpha$ are found to lead to underfitting \cite{mixup},
which could explain our observation.

\subsection{Robustness and Enhanced Recall}

To shed light on the model’s reinforced robustness, we conduct two experiments to assess its resilience to perturbations.

\begin{figure*}[h]
\centering
   \includegraphics[width=0.81\linewidth]{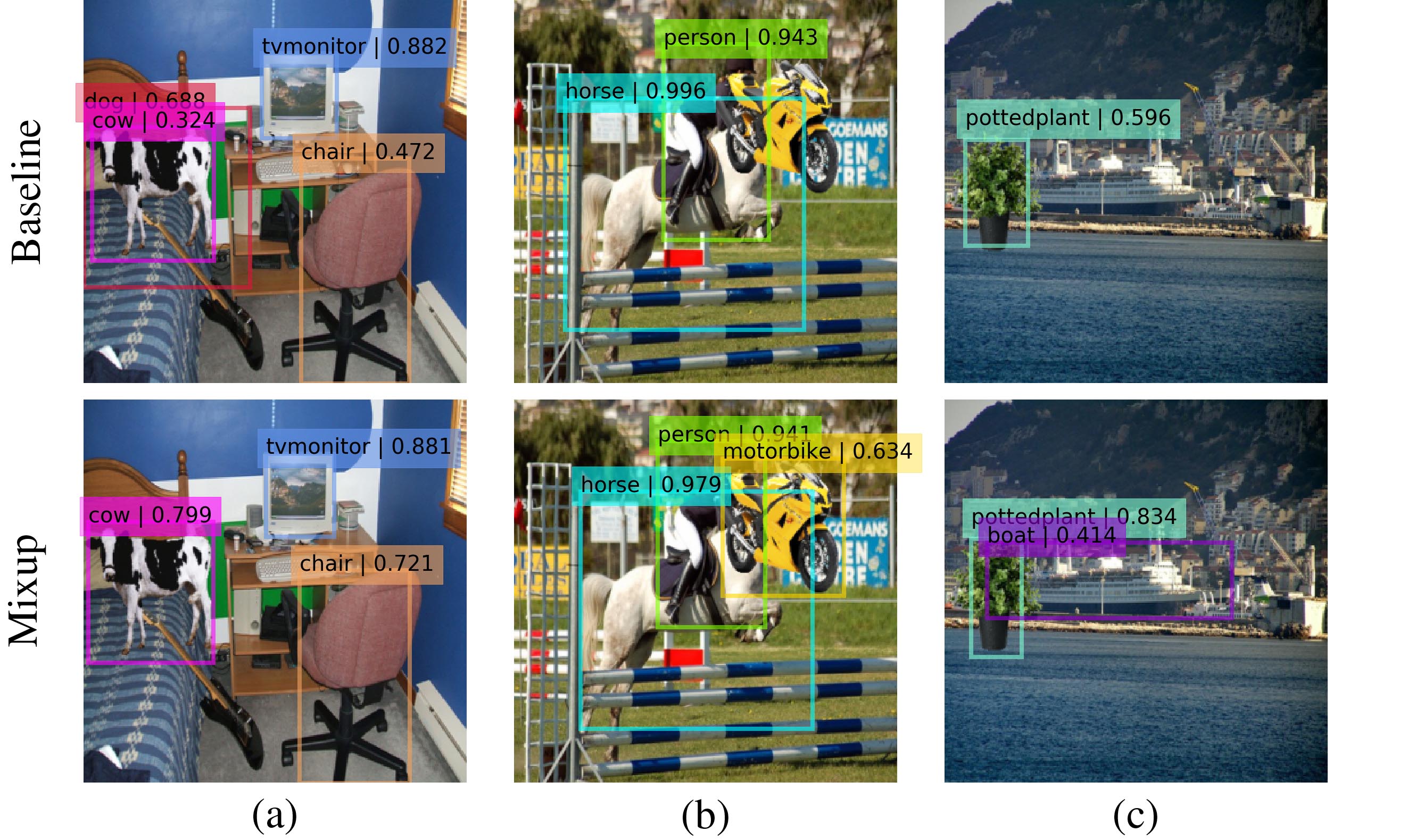}
   \caption{Examples of detection on patched images for baseline and mixup models; (a) the indoor context leads the baseline to mistake a cow for a
   dog; (b) the motorcyle is not detected due to incongruous context; (c) the patch provokes non-local alteration by discarding boat detection.}
\label{patch_detection}
\vspace*{-1em}
\end{figure*}

First, to simulate situations where the image quality is corrupted, we compare baseline and mixup models mAP when evaluated on noisy data. We set
the noise to follow a centered Gaussian distribution with standard deviation $\sigma$. The mAP is then computed in a low-
($\sigma=0.1$), intermediate- ($\sigma=0.2$) and high-corruption setting ($\sigma=0.4$). As reported in Table~\ref{noise_scores}, the
mixup model globally demonstrates a greater tolerance to noise. Interestingly, when noise intensifies, \ie we get further away from real data points,
the mixup trained model shows a greater relative performance. It suggests that mixup helps indeed with learning out of the dataset distribution.

\begin{table}[h]
\centering
\begin{tabular}{c|cccc}\toprule
$\sigma$                          & 0                       & 0.1                     & 0.2                     & 0.4                     \\\toprule
Baseline                       & 72.5\scriptsize{$\pm$0.1}                   & 70.3\scriptsize{$\pm$0.2}                    & 65.7\scriptsize{$\pm$0.6}                    & 54.3\scriptsize{$\pm$0.6}                    \\
Mixup                          & 73.1\scriptsize{$\pm$0.2}                    & 71.6\scriptsize{$\pm$0.4}                    & 68.1\scriptsize{$\pm$0.5}                    & 58.6\scriptsize{$\pm$1.2}                    \\\midrule
Difference & 0.6\scriptsize{$\pm$0.2} & 1.3\scriptsize{$\pm$0.5} & 2.4\scriptsize{$\pm$0.9} & 4.3\scriptsize{$\pm$1.7} \\\toprule
\end{tabular}
\caption{Comparison of noise tolerance in the mAP. Models are trained on VOC07+12 \texttt{trainval} and evaluated on noised VOC07 \texttt{test} data for
several noise intensity. The gap seems to grow with $\sigma$ in favor of the mixup model, suggesting that the noisier an image is, the more relevant
it is to use mixup.}
\label{noise_scores}
\vspace*{-1.6em}
\end{table}

Secondly, following the footsteps of Rosenfeld~\etal~\cite{elephant}, we design an experiment to challenge its contextual reasoning and ability to
cope with image transplantation. We collect three different open source clipped images for each of the 20 classes in PASCAL VOC by typing the class labels in a search engine and randomly selecting 3 PNG images with a transparent background. Then, for each
image in the VOC07 \texttt{test}, we successively randomly draw a patch among collected images, apply scale preserving resizing to the patch so
that it randomly occupies between 10 and 40\% of the image size, and paste it at a random location. By repeating this operation on five
independent copies of the evaluation set | summing up to a total of 24720 examples | we expect the so patched images to be on average
unrelated to their
surrounding. We then compare the baseline and mixup model on two criteria: on the one hand, detection of the transplanted patches only,
which measures the ability to predict regardless
of context. On the other hand, we also compute the overall mAP on these altered datasets, which indicates how resilient the model is to patching
induced perturbations~\cite{elephant}. Indeed, as plotted in Figure \ref{patch_detection}, we observe image transplantation has multiple effects on
detectors beyond occasional occlusion by the patch. Detectors happen to become very unstable in their predictions, with volatile confidence levels
causing patches, but also other objects in the image, to be undetected or misdetected. Besides, as supported by Rosenfeld~\etal~\cite{elephant}, these effects also occur remotely from the patch and are highly dependent on the patch nature and location.

\vspace*{-0.1em}
\begin{table}[h]
\centering
\small
\begin{tabular}{lcccc}\toprule
IoU threshold                & \multicolumn{2}{c|}{0.50} & \multicolumn{2}{c}{0.75} \\
Model                  & Baseline        & Mixup       & Baseline        & Mixup        \\\toprule
Patches Detected  & 8329            & 11693       & 5201            & 8340         \\
Patch Precision(\%)  & 76.0           & \textbf{78.3}        & \textbf{82.5}            & 79.9         \\
Patch Recall(\%)      & 33.7            & \textbf{47.4}        & 20.9            & \textbf{33.7}         \\
mAP \small{(all objects)} & 50.3            & \textbf{51.5}        & 49.1            & \textbf{50.7}         \\\toprule
\end{tabular}
\caption{Comparison of performance on random patches detection and patched images detection. Scores are averaged across classes on 5 copies
of VOC07 \texttt{test} set. Models are trained on VOC07+12 \texttt{trainval}. The IoU threshold corresponds to the bounding boxes matching criterion,
0.75 being stricter than 0.50.}
\label{patch_table}
\vspace*{-0.4em}
\end{table}

Table \ref{patch_table} shows results for default ($\tau=0.5$) and a stricter ($\tau=0.75$) matching threshold. Mixup turns out to
significantly improve patches recall, detecting roughly 50\% more patches than our baseline. It emphasizes a bolstered ability to reason
independently from context. Although we notice precision on patches detection to be mitigated when using a stricter IoU criterion, the mAP is still 
greater by a margin consistent with the one we obtain in Table~\ref{VOC_results}. It suggests an effective improvement of the detector's robustness.

\subsection{Regularization Analysis by Anchor Level}

One of the mechanisms behind the success of mixup in classification is its ability
to compress class-specific representations into manifolds with lower dimensionality~\cite{manifold_mixup}.
This is an important underlying process of generalization~\cite{DL_information, bottleneck}, referred to as \emph{flattening}. In
their experiments, Verma~\etal~\cite{manifold_mixup} take classification mixup one step further and propose to
not only interpolate samples and labels but also hidden representations. Although they point out mixup does not
always enforce class compression, their method succeeds at this task. They provide
theoretical evidence of flattening and empirically demonstrate it by computing the singular
value decomposition of class-specific hidden representations. Small singular
values for models trained with their approach were consistently reduced, resulting in
greater explained variability by the main singular component.

Inspired by this experiment, we conduct a similar study. By consenting to some degree of approximation, we
gather batches of images from VOC07 \texttt{test} according to their most frequent instance-level label, which
can be thought of as image-level labeling. We then extract image-level specific representations for baseline and
mixup models, by retrieving the penultimate classification hidden representation of the detector, \ie logits on all anchors.
As for the SSD with 300$\times$300 input size, the latter can be split into six parts $\cA_1, \ldots, \cA_6$ sorted by increasing anchor scales.
We then compute, for each image-level representation, the variance explained by the first principal component analysis (PCA) direction.
\vspace*{-0.3em}

\begin{figure}[h!]
\begin{center}
   \includegraphics[width=\linewidth]{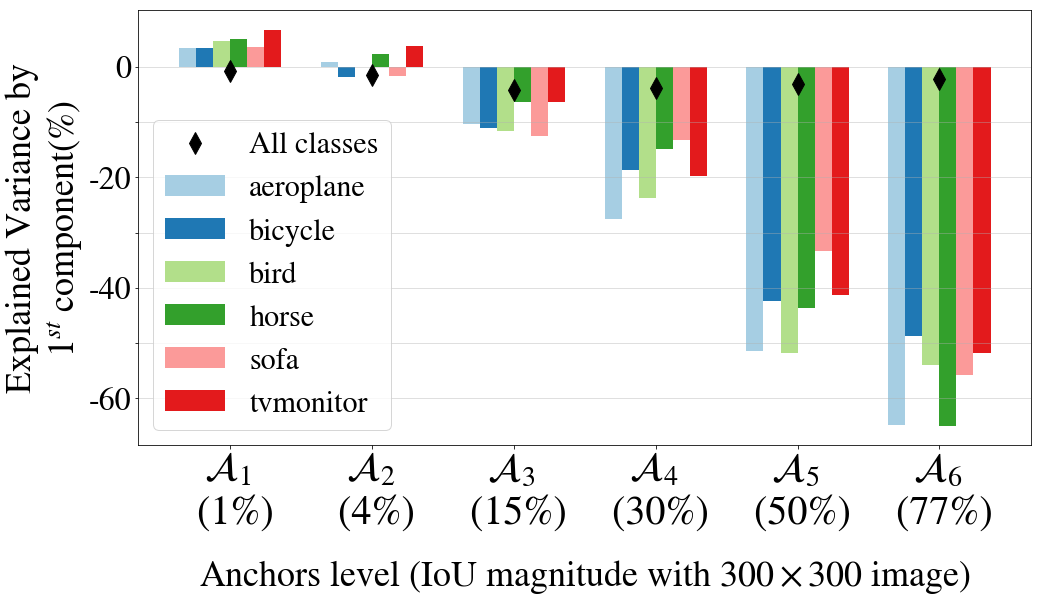}
\end{center}
   \vspace*{-0.8em}
   \caption{Difference in explained variance by the first PCA component of image-level specific hidden representation between mixup and baseline models, at multiple anchor scales; plotted for 2 vehicle (blue), animal (green) and household (red) classes. A higher bar means a flatter representation; "All classes": computed for all classes merged together. Consistently with~\cite{manifold_mixup}, the trend is only image-level
specific and not verified for all images together. It confirms this experiment highlights a class-specific phenomenon.}
\label{flattening}
\vspace*{-0.5em}
\end{figure}

Globally, mixup actually seems to scatter variability through multiple components, playing the
flattening counter game. As depicted in Figure~\ref{flattening}, this is particularly true for large anchors ($\cA_4$ to $\cA_6$) where the
information is being washed down by mixup. However, when it comes to the most granular anchor map $\cA_1$,
mixup enhances the explainability endowed to the first PCA component consistently across classes. It hence
suggests a lowered variability in image-level specific representations of small regions.

To further investigate how mixup does not benefit equally to all anchors, we train six separate models on VOC07, but each time, only use mixup on 
a unique anchor level $\cA_i\,,\, i\in\{1, \ldots, 6\}$. To truly single out the mixed anchors against all others, we emphasize the 
blending ratio and use $\alpha=0.75$ during training. At each iteration, we hence run a forward pass on both the original and the mixed batch. The 
output of the mixed batch forward pass is then used to backpropagate on $\cA_i$ logits only, while the predictions over the original images are used 
for backpropagation on all other anchor maps. Doing so, we double inference cost which causes us to reduce batchsize to 16 due to memory 
limitations. Results of Table \ref{table:branchmix_results} demonstrate that mixup indeed hurts performance when used on large 
anchors whereas scores on smaller objects detections tend to be marginally greater than the baseline for $\cA_1$ and $\cA_3$.

\begin{table}[h]
\label{table:branchmix_results}
\centering
\begin{tabular}{c}
{\normalsize Baseline mAP  \;|\; 65.2\footnotesize{$\pm$0.2} } \\
\end{tabular}

{\centering
\begin{tabular}{cc}\toprule
Mixed level & mAP  \\\midrule
$\cA_1$          & \textbf{65.9\footnotesize{$\pm$0.1}} \\
$\cA_2$          & 65.2\footnotesize{$\pm$0.5} \\
$\cA_3$          & \textbf{66.5\footnotesize{$\pm$0.3}}\\\toprule
\end{tabular}
\quad
\begin{tabular}{cc}\toprule
Mixed level & mAP  \\\midrule
$\cA_4$          & 64.7\footnotesize{$\pm$0.2} \\
$\cA_5$          & 64.6\footnotesize{$\pm$0.6} \\
$\cA_6$         & 64.8\footnotesize{$\pm$0.5} \\\toprule
\end{tabular}

\vspace{0.2em}
\caption{Evaluation on VOC07 \texttt{test} mixing for a unique anchor level; Baseline is trained without mixup and batch size 16; "Mixed level": unique anchor map level trained with mixup; left table: smaller anchors; right table: larger anchors}
\vspace*{-0.5em}}
\end{table}

Knowing that SSD main weakness 
is small objects detection~\cite{ssd}, such regularization could indeed have contributed to its bolstered performance, 
leading to the overall improved mAP we obtain in Table~\ref{VOC_results}.

\section{Discussion and Future Work}

Anchors provide us with a comprehensive representation of sets of bounding boxes, allowing us to replicate
the mixup regularization anchor-wise and further extend it to propose a novel general adaptation of Zhang~\etal algorithm
to region proposal based detectors. The proposed method stands out by enhanced robustness and
confidence in predictions at negligible additional computational costs.

We believe the proposed anchor mixing strategy could be refined to account for different blending procedures
regarding anchors sizes and better balance the overwhelming number of background anchors. Such method could
then naturally be extended to use cases such as pseudo-boxes guessing in a semi-supervised or self-supervised setting,
or help cope with low dataset size or sparse class distribution.

{\small
\bibliographystyle{ieee_fullname}
\bibliography{egpaper_for_review}
}

\end{document}